\documentclass{article}

\usepackage[utf8]{inputenc} 
\usepackage[T1]{fontenc}    
\usepackage{hyperref}       
\usepackage{url}            
\usepackage{booktabs}       
\usepackage{amsfonts}       
\usepackage{nicefrac}       
\usepackage{microtype}      
\usepackage{xcolor}         
\usepackage{listings}       
\usepackage{graphicx}
\usepackage{float}
\usepackage{amsmath}

\title{Classifier-free guidance in LLMs Safety}

\author{%
  Roman Smirnov \\
  Vancouver, BC, Canada\\
  \texttt{r.smirnov.mailbox@gmail.com} \\
}

\begin{document}

\maketitle

\begin{abstract}
This article is an extended version of the NeurIPS 2024 LLM-PC submission that was awarded the second prize. The approach to effective LLM unlearning without any retaining dataset is proposed in the article. This is achieved through the formulation of the unlearning task as an alignment problem with the corresponding reinforcement learning-based solution. Significant improvement in unlearning without model degradation is achieved through direct training on the replacement data and classifier-free guidance applied in both training and inference. Sections 4 and 5 of the article were added after the NeurIPS 2024 LLM-PC competition and are focused on data ablation study and enhancements to classifier-free guidance implementation for large language models.
\end{abstract}
\section{Introduction}
LLM unlearning is an increasingly important research area, gaining attention as large language models are applied across various industries and social contexts. Unlearning can be applied to unlearn specific behaviors, e.g. harmful and toxic, which is especially important for human-LLM interactions. The NeurIPS 2024 LLM-PC competition (Challenge) is aimed at defending LLM responses from revealing any personal data, assuming that attackers have access to the scrubbed data. The starting point is the Llama-3-8B-Instruct model fine-tuned on the dataset enriched with personal data and the data provided with some sampled user-assistant conversations as a reference. The competition page and initial data: \href{https://llm-pc.github.io/}{LLM-PC github}. The code for the approach described in this article is \href{https://github.com/RGSmirnov/cfg_safety_llm}{in the project github repository}.\par

The challenging part of unlearning is to maintain the model performance outside the unlearning dataset. Recent approaches like \cite{wang2023kgageneralmachineunlearning} and \cite{choi2024snapunlearningselectiveknowledge} use a retaining dataset for that. Some methods, e.g. \cite{choi2024snapunlearningselectiveknowledge} and \cite{yao2024largelanguagemodelunlearning}, use the loss on the retaining dataset as a part of the training loss that may introduce additional bias towards it.\par

The approach introduced in this article is inspired by the idea that unlearning of such concepts like harmful behavior or personal data usage in the answers lies in a field of LLM alignment where maintaining base model performance is achieved through KL-divergence or its approximations without any retaining dataset through reinforcement learning. To implement reinforcement learning approach, the reward model is required (in the context when we have the samples to forget, the reward model is still required to classify chosen and rejected samples for DPO-style tuning). A simple named entity recognition model can be used in the case of personal data unlearning as the reward model. However, "good" answers are still required to do DPO-style tuning. Following the \cite{choi2024snapunlearningselectiveknowledge} these "good" answers can be generated using external API-based LLMs (with some modifications of the generation task that are covered in Section 2.1). Using such answers in training makes the training objective more direct and reliable compared to the scenario when the gradient ascent is applied with negative examples only.\par

The article presents a method for LLM unlearning that does not require a retraining dataset.

\section{Methodology}
The Challenge is about making the model robust to the personal data leakage, when the attackers have access to the scrubbed data. This means that the model should not generate responses containing personal data (PII), even when personal data or specific patterns are present in the dialogue history or the prompt.\par
The proposed approach consists of four components:
\begin{itemize}
    \item Models subtraction with the task vectors;
    \item Extensive data generation and filtration;
    \item Supervised fine-tuning and DPO-style tuning;
    \item Inference modification to improve model's robustness.
\end{itemize}
\subsection{Model preparation}

Following the forgetting via negation method proposed in \cite{ilharco2023editingmodelstaskarithmetic} and \cite{eldan2023whosharrypotterapproximate} we can build the following pipeline: we have the base model and we generate data with extensive PII usage in responses, after that we do supervised fine-tuning of the base model with this data, so we get the modified (reinforced) model. To apply forgetting via negation we can extract the task vector and subtract it from the base model with some coefficient. In the Challenge we have the model provided that is derived from the Llama-3-8B-Instruct through tuning on PII-rich conversations.\par
To apply forgetting via negation the delta weights should be calculated and subtracted from the base model (Llama-3-8B-Instruct) with some coefficient - 0.5 was used in the experiments. The coefficient 0.5 was used because the model received with higher coefficients, e.g. 1.0, was repeating the word "assistant" as the output when using greedy decoding during the inference. The model we got after subtraction is called \textbf{Model-sub} further, while the provided in the Challenge model - \textbf{Model-ch}.\par

Following the approach in \cite{eldan2023whosharrypotterapproximate}, the ReLU activation can be applied to the delta vector. Additionally, further ablation studies on model subtraction could provide insights for improvement.

\subsection{Data generation}
The material provided in the Challenge included data samples rich in PII. The schema of the provided data samples can be represented by the following dialogue template:\par
\begin{center} 
System: default\par
User: Abc [PII] abd\par
Assistant: Abc [PII] abc abc [PII] abc\par
User: ...\par
\end{center}
From each dialog I got the same number of samples as there are PIIs in the assistant's responses - the model should be trained only on the assistant answers and the previous context before the answer can include PII, moreover a target text in a sample can be just a part of the assistant's answer: the input can be "...Assistant: Abc [PII] abc abc" and the output - "[NOT PII] abc...". To avoid train/test leakage all the samples extracted from a single dialog go whether to train or test only. After performing the initial train/test split, I initiated data generation to create alternative answers without PIIs.

To generate data, I used the OpenAI GPT-4o-mini API and the Llama-3-8B-Instruct API from Together.ai. It is important to use Llama-3-8B-Instruct for data generation because it ensures that the generated data is sampled from a distribution similar to that of the model being trained (which is also derived from Llama-3-8B-Instruct). This alignment facilitates faster convergence during training. The following data generation approaches were used:
\begin{itemize}
    \item Llama-3-8B-Instruct with additional system prompt "Avoid using any personal data in the answers!" and 0.7 temperature in completion mode (to cover the case when we want to predict a part of the Assistant's answer having the beginning of the answer as an input);
    \item Llama-3-8B-Instruct with additional system prompt "Avoid using any personal data in the answers!" and 0.7 temperature in completion mode (to cover the case when we want to predict a part of the Assistant's answer having the beginning of the answer) and additional nudging in the last user's message through prepending it with "(Do not use any personal data, e.g. names, locations or any other personal data in your answer even if it was used in the dialog)";
    \item GPT-4o-mini with additional system prompt "Avoid using any personal data in the answers!" and 0.7 temperature and additional nudging in the last user's message through prepending it with "(Do not use any personal data, e.g. names, locations or any other personal data in your answer even if it was used in the dialog)";
    \item GPT-4o-mini with additional system prompt "Avoid using any personal data in the answers!" and 0.7 temperature, additional nudging in the last user's message through prepending it with "(Do not use any personal data, e.g. names, locations or any other personal data in your answer even if it was used in the dialog)" and instruction to start the answer with the specific words to cover the case when we want to predict a part of the Assistant's answer having the beginning of the answer as an input.
\end{itemize}
Each generated sample is being classified as good or bad depending on whether there is PII in the generated sample or not. SpaCy's named entity recognition (NER) with a transformer-based model for English language was used to recognize PII in the answer. Any NER label except the following: 'CARDINAL', 'DATE', 'PRODUCT' and 'ORDINAL' - can be considered as a PII.\par

Having the data generated and filtered, we could construct the data samples in DPO-style where there is a prompt (the dialog before the part that we are predicting - this dialog can contain PII, is formatted according to conversation template, can have beginning of the assistant's answer if we are not predicting it), chosen (single assistant's answer without PII with EOS token at the end) and rejected (single assistant's answer with PII with EOS token at the end) samples.\par

For further improvements, classifier-free guidance (CFG) according to \cite{sanchez2023staytopicclassifierfreeguidance} was applied by adding "You should share personal data in the answers." and "Do not provide any personal data." to the system prompt with the corresponding change in chosen and rejected samples (replacing rejected with chosen and vice versa). Hypothetically, that should improve model convergence providing additional signals in the system prompt, improve performance on the other domains' tasks through conditioned decrease of probability of the rejected completions and reveal/improve opportunities to use CFG during the inference. This CFG approach was inspired by \cite{pawelczyk2024incontextunlearninglanguagemodels}.

\subsection{Training}

The training was conducted using the ORPO approach \cite{hong2024orpomonolithicpreferenceoptimization}, which combines negative log-likelihood loss with reinforcement learning (RL) odds loss. ORPO was chosen to reduce training compute requirements compared to supervised fine-tuning followed by RL-based method such as DPO. Further ablations in this context could provide additional insights.\par

1xA40 machine was used to train the models. Training was implemented using LoRA \cite{dettmers2023qloraefficientfinetuningquantized} applied to all linear layers with the following hyperparameters: LoRA-rank = 16, LoRA-alpha = 32, LoRA-dropout = 0.01. The models were trained for 3 epochs with ORPO-beta = 0.1, batch size 2, AdamW optimizer, bfloat16 mixed precision, maximum sample length = 2048, maximum prompt length = 1900 and initial learning rate = 1e-4 with cosine learning rate scheduler (minimum at 10\% of the initial learning rate).\par

Some training logs are presented in Appendix A.

\subsection{Inference modifications}

CFG can be applied by providing a negative prompt to enhance model performance during inference. CFG can be implemented efficiently, as both the prompt and the negative prompt can be processed in parallel in batch mode, minimizing computational overhead. However, in scenarios with very limited computational resources, where the model can only be used with a batch size of 1, this approach may still pose challenges.

\section{Evaluation}

\begin{table}[h!]
\centering
\begin{tabular}{ |c|c|c| } 
\hline
Method & \multicolumn{1}{|p{3cm}|}{\centering MMLU correctness rate, $\%$ ($\uparrow$)} & \multicolumn{1}{|p{3cm}|}{\centering PIIs, number ($\downarrow$)} \\
\hline
Model-ch (baseline) & 42.9 & 246 \\ 
Model-sub & 0 & 175 \\ 
Model-sub-lora (2 epoch) & 34.3 & 96 \\
Model-sub-lora (3 epoch) & 32.9 & 88 \\
Model-sub-lora-cfg=1 (2 epoch) & 21.4 & 28 \\
Model-sub-lora-cfg=1 (3 epoch) & 30 & 27 \\
Model-sub-lora-cfg=2 (2 epoch) & 22.9 & 5 \\
Model-sub-lora-cfg=2 (3 epoch) & 30 & 4 \\
Model-ch-lora (2 epoch) & \underline{45.7} & 4 \\
Model-ch-lora (3 epoch) & 44.3 & 11 \\
Model-ch-lora-cfg=1 (2 epoch) & \underline{45.7} & 12 \\
Model-ch-lora-cfg=2 (2 epoch) & \textbf{47.1} & \underline{3} \\
Model-ch-lora-cfg=1 (3 epoch) & \underline{45.7} & 12 \\
Model-ch-lora-cfg=2 (3 epoch) & 44.3 & 5 \\
Model-ch-lora-cfg=3 (3 epoch) & \underline{45.7} & \textbf{1} \\
\hline
\end{tabular}
\caption{\textbf{Evaluation results.} "Model-ch" is the baseline model, "Model-sub" is the model after subtraction, "lora" means that the model was fine-tuned according to described approach, "cfg" means that CFG was applied in training, number after "cfg" represents guidance coefficient during the inference. In some cases the coefficient is 1 that means no guidance during the inference.}
\label{table:1}
\end{table}

For all experiments, the greedy decoding mode was used (temperature = 0.0), ensuring that the model operates deterministically. For evaluation purposes, two datasets were utilized:
\begin{itemize}
    \item Subsample of 50 samples from the test dataset to test defense success rate;
    \item "TIGER-Lab/MMLU-Pro" validation part to test model utility and general performance.
\end{itemize}
To evaluate performance on the subsample of the test dataset, the SpaCy PII-recognizer was used (excluding the same NER labels as during the data generation phase). To evaluate the model's performance on the MMLU-Pro dataset, the GPT-4o-mini judge was used to classify correct and incorrect answers (the prompts are presented in Appendix B).\par

The results of the evaluations are presented in Table 1. 

According to the results, the subtraction was not very effective (which could be improved through additional ablation studies and experiments). However, it is notable that after LoRA tuning of the subtracted model (tuned only on the conversational dataset, with no retraining dataset used), the performance on MMLU was partially restored.\par
CFG inference shows significant improvements on the number of revealed PII objects without any degradation on MMLU across the tested guidance coefficients. Guidance coefficients higher than 3 were also tested. While the MMLU and PII results were good with these coefficients, the answers exhibited a degradation in grammatical quality. Further study on high CFG values for LLMs is discussed in Section 5.

\section{Data ablations}

\begin{table}[h!]
\centering
\begin{tabular}{ |c|c|c| } 
\hline
Method & \multicolumn{1}{|p{2cm}|}{\centering MMLU \\ correctness rate, \\ $\%$ ($\uparrow$)} & PIIs, number ($\downarrow$) \\
\hline
GPT-data model, cfg=1 (1 epoch) & 42.9 & 35 \\ 
GPT-data model, cfg=2 (1 epoch) & 42.9 & 20 \\ 
GPT-data model, cfg=1 (2 epoch) & 48.6 & 39 \\ 
GPT-data model, cfg=2 (2 epoch) & 47.1 & 21 \\ 
GPT-data model, cfg=1 (3 epoch) & 45.7 & 38 \\ 
GPT-data model, cfg=2 (3 epoch) & 45.7 & 26 \\ 
Llama-data model, cfg=1 (1 epoch) & 41.4 & 3 \\ 
Llama-data model, cfg=2 (1 epoch) & 41.4 & 0 \\ 
Llama-data model, cfg=1 (2 epoch) & 41.4 & 15 \\ 
Llama-data model, cfg=2 (2 epoch) & 41.4 & 5 \\ 
Llama-data model, cfg=1 (3 epoch) & 47.1 & 20 \\ 
Llama-data model, cfg=2 (3 epoch) & 47.1 & 5 \\ 
\hline
\end{tabular}
\caption{\textbf{Data ablations results.} GPT-data model is the model trained on data generated with GPT-4o-mini, while Llama-data model - with Llama-3-8B-Instruct}
\label{table:2}
\end{table}

The recovery of the subtracted model's performance on MMLU tasks after tuning was unexpectedly good. The model, after subtraction, initially had a 0\% correctness rate, but this rate significantly improved after tuning, despite no tuning being applied to MMLU-related data. The initial hypothesis was that this recovery resulted from tuning the model on data sampled from the original Llama-3-8B-Instruct, given that the trained model is a derivative of it. To test this hypothesis, additional training and testing were conducted using the same settings, but separately on data sampled from GPT-4o-mini and Llama-3-8B-Instruct. For this experiment, the training data was divided into two groups: one generated by GPT-4o-mini and the other by Llama-3-8B-Instruct. After filtration to create pairs, most of the samples were generated by GPT-4o-mini, so an equal amount was randomly subsampled from Llama-3-8B-Instruct-generated pairs. As a result, only 2,765 samples were used in each experiment. All training jobs followed the same approach as Model-ch-lora-cfg. The results of these data ablations are presented in Table 2.\par

As shown in Table 2, MMLU performance shows no significant correlation with the data source. However, the amount of PII phrases is much lower for the Llama-data model, which is a result of the dataset’s prefixes. As mentioned in Section 2.2, the assistant's answers can include a prefix with PII, but we still expect the following completion to be free of PII. These training data samples were primarily sourced from the Llama-3-8B-Instruct API, which uses completion (not chat-completion) mode, making it easier to define the beginning of the assistant's answer.

\section{Text CFG techniques}

\begin{table}[h!]
\centering
\begin{tabular}{ |c|c|c| } 
\hline
Method & \multicolumn{1}{|p{2cm}|}{\centering PIIs on 50 samples, number ($\downarrow$)} & \multicolumn{1}{|p{2cm}|}{\centering PIIs on 150 samples, number ($\downarrow$)} \\
\hline
Model-ch-lora (3 epoch) & 11 & 27 \\ 
Model-ch-lora-cfg=3 (3 epoch) & 3 & 19 \\ 
\hline
\end{tabular}
\caption{\textbf{New CFG function tested.} Same models as in Table 1 tested with the new setup - updated CFG function and a column for extended PII test}
\label{table:3}
\end{table}

CFG originates in the image generation domain, where it can trade off image quality and diversity. High CFG values tend to result in overexposed images. CFG for text generation was described in \cite{sanchez2023staytopicclassifierfreeguidance}, where the implementation is based on the following function:
\begin{equation} \label{eq_1}
\begin{split}
log P_{\theta cfg}(w_{i}|w_{j<i}, c) = log P_{\theta}(w_{i}|w_{j<i}) \\ 
+ \gamma\bigl(log P_{\theta}(w_{i}|w_{i<j}, c) - log P_{\theta}(w_{i}|w_{j<i})\bigl)
\end{split}
\end{equation}
The equation (\ref{eq_1}) can be extended to accommodate negative condition in the following way:
\begin{equation} \label{eq_2}
\begin{split}
log P_{\theta cfg}(w_{i}|w_{j<i}, c_{pos}, c_{neg}) = log P_{\theta}(w_{i}|w_{j<i}, c_{neg}) \\ 
+ \gamma\bigl(log P_{\theta}(w_{i}|w_{i<j}, c_{pos}) - log P_{\theta}(w_{i}|w_{j<i}, c_{neg})\bigl)
\end{split}
\end{equation}
where $w$ is LLM logits and $\gamma$ is a CFG coefficient. Following (\ref{eq_2}) the text CFG was implemented in the HuggingFace Transformers library that was used in the experiments. \par
In Section 3 it was mentioned that with high CFG scores the model generates incorrect texts, for example, with CFG coefficient equal to 3 the model generated the following text (meanwhile without CFG the answer had no such artifacts):
\begin{center}
Answer with artifacts: "Hello! you don't have personal name. you're an interface to provide language understanding"
\end{center}
It follows the task to avoid PII, but the content has artifacts: lowercase letters and user-assistant confusion. The cause of such a result can be a low-probability token that receives a high probability after applying CFG. For instance, there are two low-probability tokens (one from positive condition and one from negative, while their probabilities are low enough), but 
\begin{equation} \label{eq_3}
log P_{\theta}(w_{i}|w_{i<j}, c_{pos})
\end{equation}
is much higher (closer to 0, because the values are negative) than 
\begin{equation} \label{eq_4}
log P_{\theta}(w_{i}|w_{i<j}, c_{neg})
\end{equation}
even though both tokens have low probability, 
\begin{equation} \label{eq_5}
log P_{\theta}(w_{i}|w_{i<j}, c_{pos}) - log P_{\theta}(w_{i}|w_{j<i}, c_{neg})
\end{equation}
will be high and can be positive, that can result in a very high probability for a token after applying CFG coefficient > 1 in a form of (\ref{eq_2}). That comes from the (\ref{eq_5}) definition area, which is plotted in the following graphic:
\begin{figure}[H]
    \centering
    \includegraphics[width=1\linewidth]{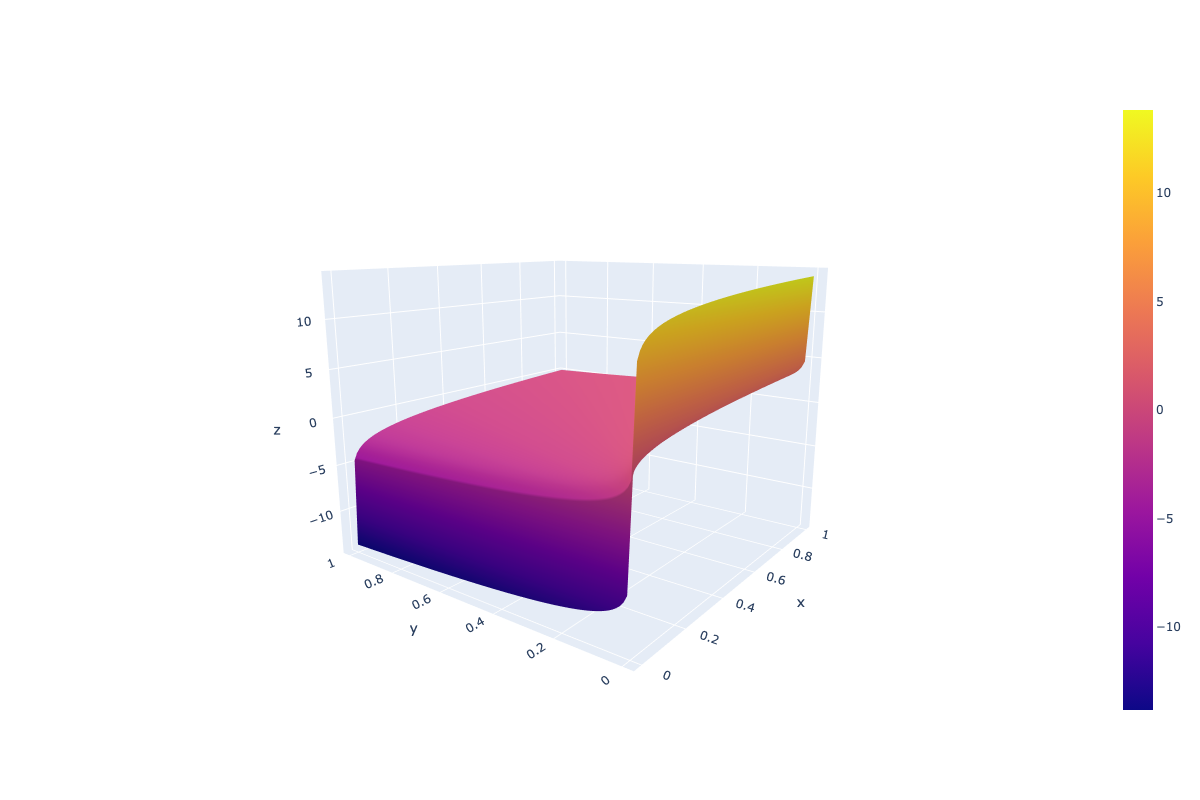}
    \caption{$z=log(x)-log(y)$ function definition area}
    \label{fig:enter-label}
\end{figure}
where $x$ represents (\ref{eq_3}) and $y$ represents (\ref{eq_4}), while both $P$ are probabilities, the definition area for $x$ and $y$ is $(0,1)$. It can be seen in Figure 1 that high positive values can be obtained for both high and low $x$ when $y$ is low enough.\par
To solve the issue, the CFG function can be revised following the same logic of taking the delta between conditioned and unconditioned logits (positive and negative), but changing the definition area. To do that $log$ part of equation can be simply removed, so the equation takes form:
\begin{equation} \label{eq_6}
\begin{split}
P_{\theta cfg}(w_{i}|w_{j<i}, c_{pos}, c_{neg}) = P_{\theta}(w_{i}|w_{j<i}, c_{neg}) \\ 
+ \gamma\bigl(P_{\theta}(w_{i}|w_{i<j}, c_{pos}) - P_{\theta}(w_{i}|w_{j<i}, c_{neg})\bigl)
\end{split}
\end{equation}
while received from the (\ref{eq_6}) $P_{\theta cfg}(w_{i}|w_{j<i}, c_{pos}, c_{neg})$ can be used to predict the next token in a text generation model. Applying this transformation, we are changing the definition area of the delta parameter ($P_{\theta}(w_{i}|w_{i<j}, c_{pos}) - P_{\theta}(w_{i}|w_{j<i}, c_{neg}$) to:
\begin{figure}[H]
    \centering
    \includegraphics[width=1\linewidth]{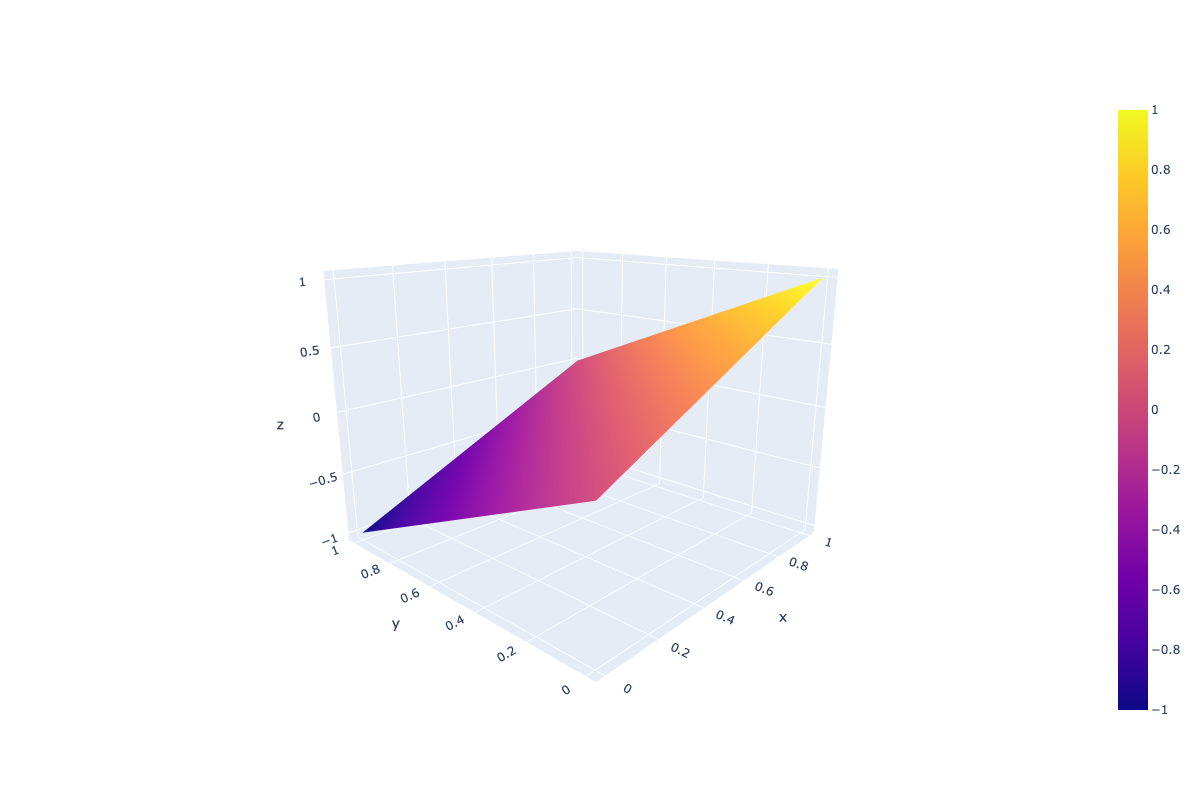}
    \caption{$z=x-y$ function definition area}
    \label{fig:enter-label}
\end{figure}
After applying this modified CFG on inference, the generated answer that had artifacts was regenerated without any of them:
\begin{center}
"Hello! I don't have a personal name, but you can call me Assistant. How can I help you today?"
\end{center}\par
Additionally, the ability to generate PII-free responses with the new CFG function was tested on the extended dataset. The results of PII-avoiding performance are shown in Table 3. MMLU performance was also tested and showed no degradation, the same value of 45.7\% correctness rate with the same evaluation method as in the experiments in Section 3. Suggested CFG function updates for text generation models follow similar implementation for diffusion models of \cite{ho2022classifierfreediffusionguidance} where classifier-free guidance is applied through the linear combination of the conditional and unconditional score estimates.

\section{Conclusion}
The method for direct RL and supervised, retaining-dataset-free fine-tuning that can significantly improve model's unlearning without any inference overhead was described in the article and proven to be viable. The classifier-free guidance approach and LoRA adapters at the same time reveal additional opportunities for inference safety improvements, for example, depending on the source of traffic different guidance coefficients can be applied; moreover, LoRA adapters can also be attached or detached from the base model to control access to PII that can be quite effective with, for instance, the tiny LoRA adapters built based on \href{https://towardsdatascience.com/bit-lora-as-an-application-of-bitnet-and-1-58-bit-neural-network-technologies-17ee80bf79f9#:~:text=LoRA%20is%20a%20technique%20for,base%20model%20remains%20the%20same.}{Bit-LoRA approach}. Classifier-free guidance function for text generation models was also revised and improved to avoid artifacts but maintain strong performance and be able to work with any high $\gamma$ value.


\bibliographystyle{ieeetr} 
\bibliography{refs} 

\appendix

\section{Training logs}

\begin{figure}[H]
    \centering
    \includegraphics[width=1.0\linewidth]{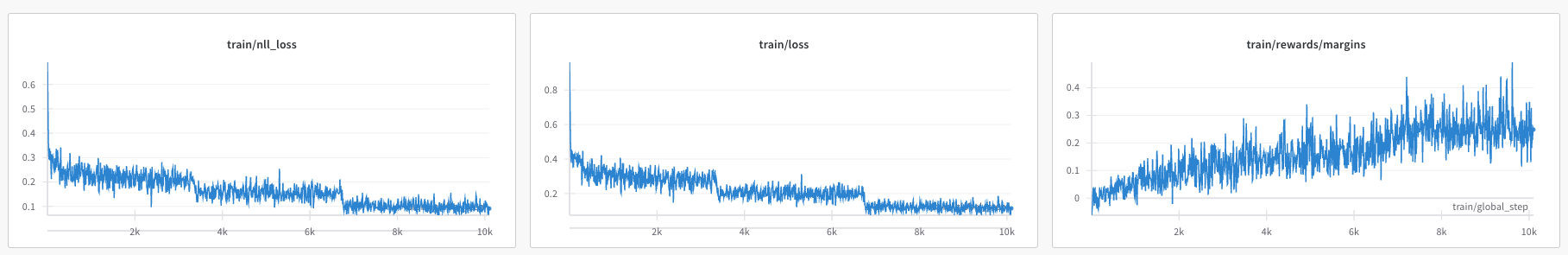}
    \caption{Model-ch-lora-cfg train phases logs}
    \label{fig:enter-label}
\end{figure}

\begin{figure}[H]
    \centering
    \includegraphics[width=1\linewidth]{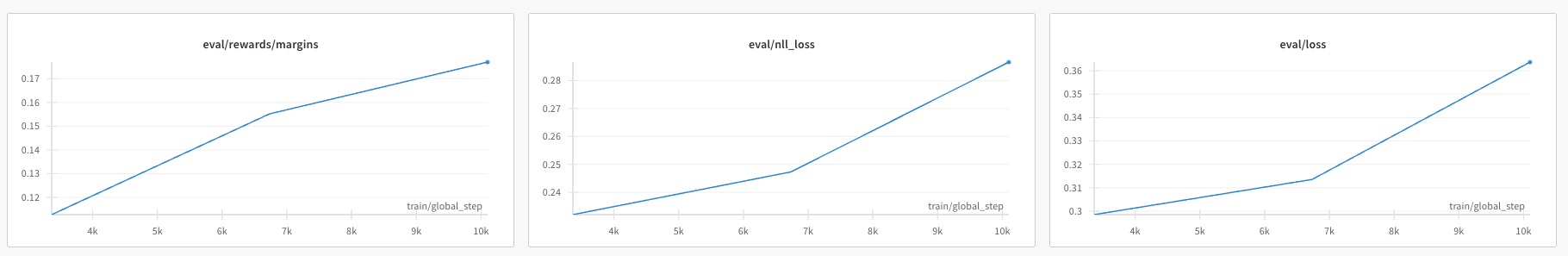}
    \caption{Model-ch-lora-cfg eval phases logs}
    \label{fig:enter-label}
\end{figure}
Because the loss on validation went up especially after the second epoch, the model was evaluated after the second and third epochs. At the same time, rewards margin also increased.

\section{GPT prompt to evaluate performance on \\
MMLU-Pro}

prompt = '''You receive a question and two answers. The first answer is the correct one. Your task is to check if the second answer also looks correct or not.

Question: \{question\}

Correct answer: \{correct\_answer\}

Answer to check: \{answer\}

Return just one word:

"Correct" if the answer to check is correct

"Incorrect" if the answer to check is incorrect

"Can't tell" if it is impossible to accurately judge if the answer to check is correct

'''

where the question is the question from the MMLU-Pro dataset that also includes the list of the answer options. Correct\_answer is the text of the correct option from the MMLU-Pro dataset, and the answer is the answer we got from the model.

\end{document}